\documentclass[runningheads]{llncs}
\usepackage{graphicx}
\usepackage{amsmath,amssymb} 
\usepackage{color}
\usepackage{xcolor}
\usepackage{multirow}
\usepackage{hyperref}
\usepackage{subfigure}
\usepackage{booktabs}

\hypersetup{colorlinks,linkcolor={red},citecolor={green},urlcolor={red}}
\begin{document}

\newcommand*\rfrac[2]{{}^{#1}\!/_{#2}}
\newcommand{\layername}[1]{{\fontfamily{qcr}\selectfont#1}}
\newcommand{\myparagraph}[1]{{\vspace{0.5em} \noindent \bf #1}}

\pagestyle{headings}
\mainmatter

\def\eg{\emph{e.g.}} \def\Eg{\emph{E.g.}}
\def\ie{\emph{i.e.}} \def\Ie{\emph{I.e}\onedot}
\def\cf{\emph{c.f.}} \def\Cf{\emph{C.f}\onedot}
\def\etc{\emph{etc}.} \def\vs{\emph{vs}\onedot}
\def\wrt{w.r.t.} \def\dof{d.o.f.}
\def\etal{\emph{et al.}}

\title{Oriented Objects as pairs of Middle Lines} 
%

\titlerunning{Oriented Objects as pairs of Middle Lines}

\authorrunning{Haoran Wei, Yue Zhang. et al.}

\author{Haoran Wei\inst{1,2,3}, Yue Zhang\inst{1,3}, Zhonghan Chang\inst{1,2,3},  Hao Li\inst{1,3}, Hongqi Wang\inst{1,2,3} and Xian Sun\inst{1,3,*}}


\institute{\quad ${}^1$Aerospace Information Research Institute, Chinese Academy of Sciences, Beijing 100190, China.\quad ${}^2$ School of Electronic, Electrical and Communication Engineering, University of Chinese Academy of Sciences, Beijing 100190, China.\quad ${}^3$ Key Laboratory of Network Information System Technology (NIST), Aerospace Information Research Institute, Chinese Academy of Sciences, Beijing 100190, China. 
\email{weihaoran18@mails.ucas.ac.cn}
}
\maketitle

\begin{abstract}

The detection of oriented objects is frequently appeared in the field of natural scene text detection as well as object detection in aerial images. Traditional detectors for oriented objects are common to rotate anchors on the basis of the RCNN frameworks, which will multiple the number of anchors with a variety of angles, coupled with rotating NMS algorithm, the computational complexities of these models are greatly increased. In this paper, we propose a novel model named Oriented Objects Detection Network ($O^{2}$\textit{-DNet}) to detect oriented objects by predicting a pair of middle lines inside each target. $O^{2}$\textit{-DNet} is an one-stage, anchor-free and NMS-free model. The target line segments of our model are defined as two corresponding middle lines of original rotating bounding box annotations which can be transformed directly instead of additional manual tagging. Experiments show that our $O^{2}$\textit{-DNet} achieves excellent performance on \textit{ICDAR 2015} and \textit{DOTA} datasets. It is noteworthy that the objects in \textit{COCO} can be regard as a special form of oriented objects with an angle of 90 degrees. $O^{2}$\textit{-DNet} can still achieve competitive results in these general natural object detection datasets.

\end{abstract}

\section{Introduction}

\begin{figure*}[t]
	\centering
	\includegraphics[width=12cm]{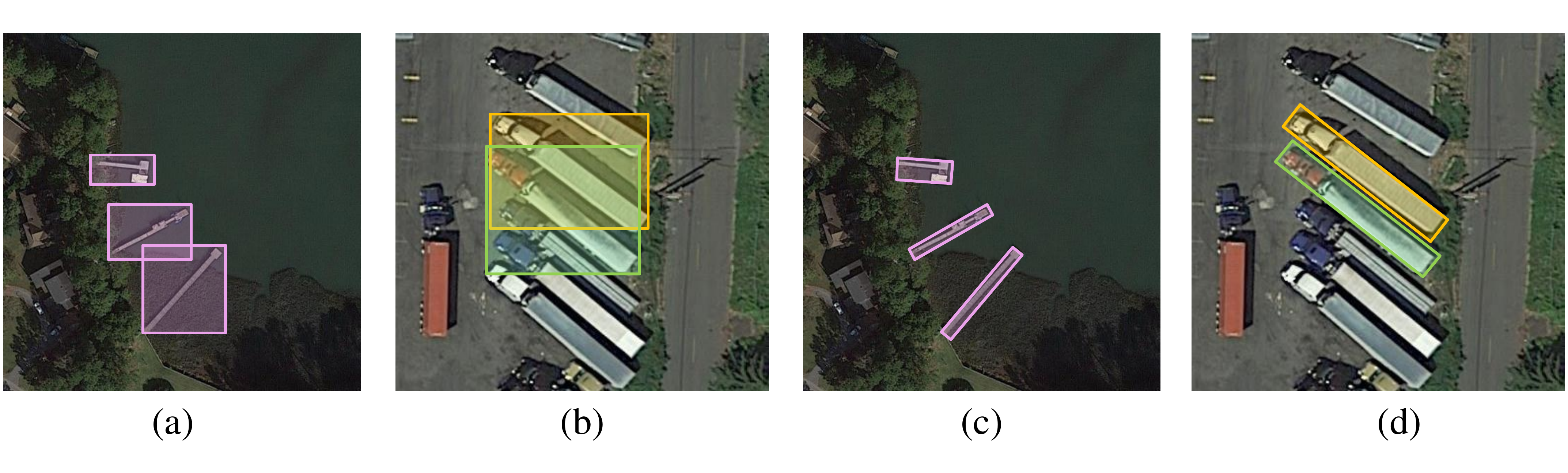}
	\caption{Horizontal bounding box \textbf{vs.} Oriented bounding box. Compared with Figure (c), the bounding box in Figure(a) carries too much redundant information. The trucks in Figure (b) and (d) have a large aspect ratio and park densely. When detecting them with horizontal bounding box, NMS algorithm will lead to missed detection due to the intersection-over-union (IOU) of two objects is too large. This problem can be solved via introducing oriented bounding box like Figure (d).}
	\label{Figure 1}
\end{figure*}

When the object such as text in natural scene and aerial target(e.g., airplane, ship, vehicle) appear in an image with a certain degree, the output form of horizontal bounding box which usually used in the detection of natural object no longer meet the detection requirement generally for it may include many redundant pixels which belong to background actually. Moreover, when detecting some objects which have a large aspect ratio and park densely, the cooperation between horizontal bounding box and NMS is easy to cause missed detection as shown in Fig. \ref{Figure 1}. In order to solve problems aforementioned, an oriented bounding box output form has been proposed, and the following detection of oriented objects has attracted more and more attention in recent years.

Many of recent research achievements of oriented objects detection rely on RCNN frameworks heavily. In the field of text detection in natural scene images, R2CNN\cite{jiang2017r2cnn} adds a new branch to regress two points and the height of oriented bounding box based on Faster R-CNN\cite{ren2015faster}. However, during the training stage, horizontal anchors are still be suppressed by NMS within the RPN network when the objects need to be detected have large aspect ratios and park densely. RRPN\cite{ma2018arbitrary} proposes rotation anchors to replace the horizontal ones and the corresponding NMS is also replaced by a new NMS algorithm calculated by rotating IOU. However, in order to cover objects of any angles, more anchors with different angles, aspect ratios and scales need to be set which increases the computational complexity of the model. Moreover, the introduction of rotation anchors in RRPN adds additional regression of angle information via Smooth L1\cite{girshick2015fast}, but the combination of rotating IOU and Smooth L1 is not perfect because the angle has boundary problem, so the oriented bounding box output by this algorithm is not accurate and often accompanied by angle jitter.

In the field of aerial images, the detection of oriented object is more difficult campare with text detection in natural scene for the complex background as well as the variation of spatial resolution of images.  SCRDet\cite{yang2019scrdet} proposes an IOU Loss to address the boundary problem for oriented bounding box mentioned in RRPN. In addition, a Multi-Dimensional Attention Network is added to deal with the complex background of aerial images. However, SCRDet is still anchor-based and NMS-based and also faces some problems bring by them. For instance, in the testing stage, 300 proposals are taken from 10000 regression boxes by NMS in \cite{yang2019scrdet} but according to our statistics, for DOTA dataset, a crop image of $800 \times 800$ size can reach up to $2000 +$ objects, which will cause missed detection.

\begin{figure*}[t]
	\centering
	\includegraphics[width=12.3cm]{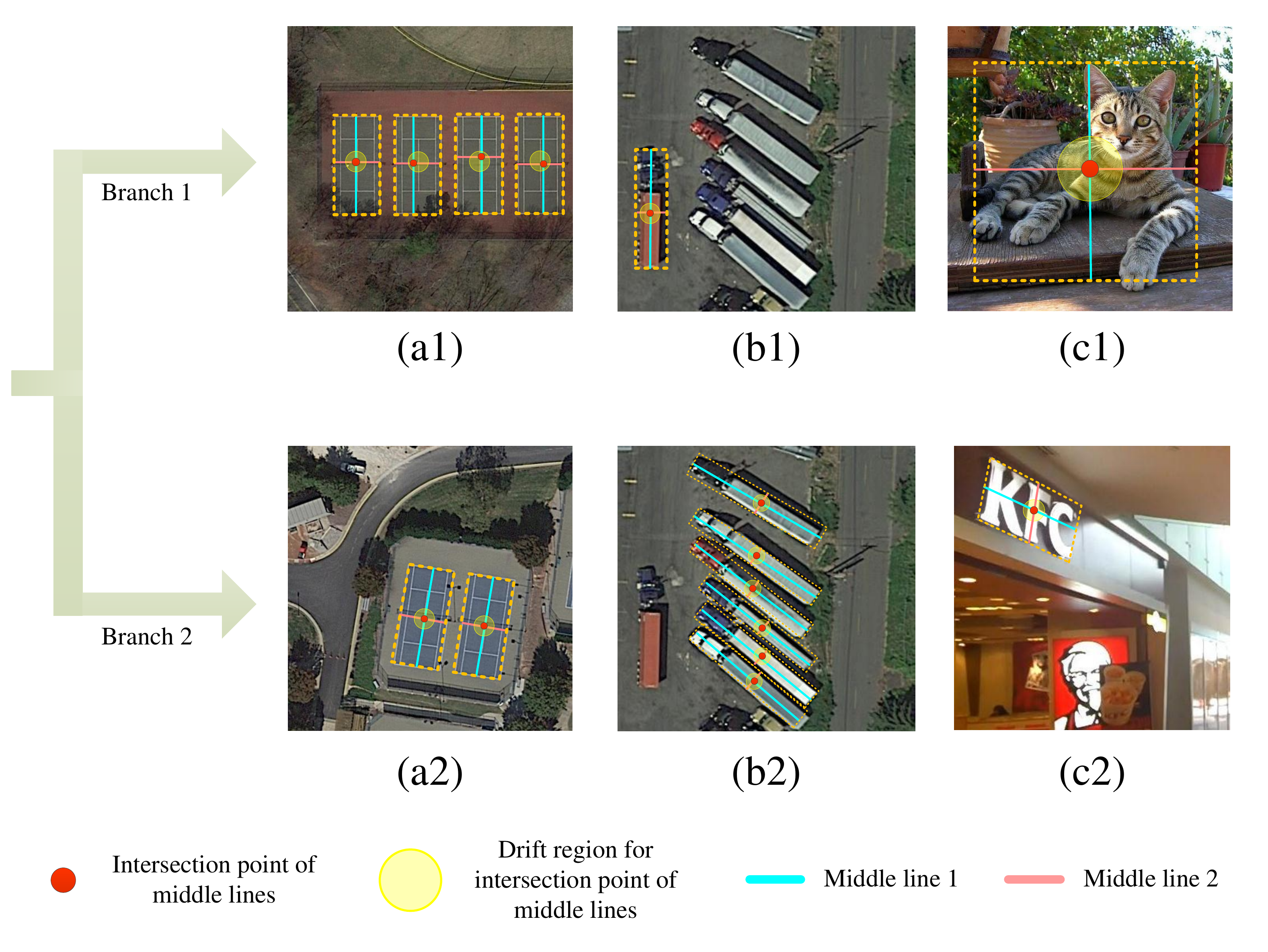}
	\caption{Views of $O^{2}$\textit{-DNet}. Figure (a1) shows that if the intersection point drift in the drift region, the final bounding box do not drift. If the angle of oriented object is 90, they will be predicted by branch 1, and objects with other angles will be detected by branch 2. In ICDAR 2015, the shape of bounding box of text is not always a rectangular, so we set the vertical loss in Line Loss to $0$ when detecting text in natural scene.} 
	\label{Figure 2}
\end{figure*}

Recently, in view of the disadvantages of the anchor-based models, a number of anchor-free algorithms\cite{law2018cornernet,zhou2019bottom,tian2019fcos,zhou2019objects} have emerged. CornerNet\cite{law2018cornernet} and ExtremeNet\cite{zhou2019bottom} can be divided into the process of detection and grouping of keypoints. While it is not suitable to apply them into aerial images which may obtain many objects due to the high time complexity of their grouping algorithm.  CenterNet\cite{zhou2019objects} puts forward a new method of regressing the height and weight of object at the center point. In order to achieve NMS-free, they obtain the center point in heatmap through a method of searching 8-connect of each center point. However, this method still needs to choose $K$ top scoring objects which can not be applied to single image with numerous targets. In addition, it may lead to the drift of the center point, and further cause the bounding box to drift. Different from the anchor-free models based on keypoints detection above, FCOS\cite{tian2019fcos} belongs to the category of dense sampling, which regress in numerous pixels for one objects. Because the bounding box of an object is regressed by a large number of pixels, NMS is still needed to filter the redundant boxes in FCOS.

In this paper, we propose a novel anchor-free and NMS-free model named $O^{2}$\textit{-DNet} as shown in Fig. \ref{Figure 2} to detect oriented objects by a pair of middle lines. Our model is a new form of anchor-free which combines the mothods of keypoints detection and dense sampling.  In order to reach the aim of NMS-free, we choose the method of keypoint detection to locate the intersection point of each pair of median lines. For the problem of intersection point drift, we design a drift region inspired by the method of dense sampling to ensure that the intersection point drift in the drift region will not affect the position of the final bounding box.  In order to successfully predict the middle line, we design a specific Line Loss according to the characteristics of lines(e.g., length, slope, position) to regress each median line. The Line Loss consists of three parts: the position loss to control the location of the endpoints of each middle line, the parallel loss to control the two endpoints of each middle line and the intersection point of two middle lines are collinear. The last one is vertical loss to control the geometric relationship between two median lines of one object. There is an order for the regression of endpoints of middle lines, so we wil also face the boundary problem. In order to solve it, we design $O^{2}$\textit{-DNet} as two branches, one for predicting horizontal objects with 90 degrees and the other for oriented objects of other angles. The design of two branches also enables us to apply $O^{2}$\textit{-DNet} to COCO\cite{lin2014microsoft}  without changing any network structure.

Our contributions and innovations are as follows:

(1) We propose a noval anchor-free and NMS-free model named $O^{2}$\textit{-DNet} to detect oriented objects by a pair of middle lines.

(2) Our $O^{2}$\textit{-DNet} is a new form of anchor-free which combines keypoints detection and dense sampling, and we design drift region to relax the requirement for accurate extracting keypoints. 

(3) Our $O^{2}$\textit{-DNet} can detecting oriented objects and horizontal objects under a single network without increasing computational complexities via two branches. For the regression of middle lines, we design a special Line Loss. 

The rest of this paper is organized as follows: In  {\color{red}\textit{Section \ref{section:2}}}, we introduce the related works done by researchers before and basic principle in our method. The details of our network and algorithms are shown in {\color{red}\textit{Section \ref{section:3}}}. We place our experiments results and analysis in {\color{red}\textit{Section \ref{section:4}}}. At last, our work is summarized and concluded in {\color{red}\textit{Section \ref{section:5}}}.

\section{Related Works} \label{section:2}

\subsection{Detectors based on Manually Engineered Features}

Traditional object detectors mainly depend on manually engineered features, they first select features like Histogram of Oriented Gradient (HOG)\cite{dalal2005histograms} generally and then input them to classifier such as Support Vector Machine (SVM)\cite{cortes1995support} to identify the existence of object.  The generalization capability of these detectors is limited by features extraction and the robustness of this type of detector needs to be further improved. 

\subsection{Detectors based on DCNNs}

In recent years, the success and development of deep convolution neural networks (DCNNs)\cite{lecun1998gradient,krizhevsky2012imagenet} bring great progress to the field of object detection. Compared with tradition detectors aforementioned, detectors based on DCNNs \cite{ren2015faster,law2018cornernet,zhou2019bottom,tian2019fcos,zhou2019objects,redmon2016you,redmon2017yolo9000,liu2016ssd} can automatically extract features through the backbone networks\cite{simonyan2014very,he2016deep}, and the accuracy as well as robustness of models is greatly improved. There are two branches which are anchor-based and anchor-free in DCNNs based detectors at present.

\subsection{Anchor-Based vs. Anchor-Free Detectors}

The concept of anchor was proposed in Region Proposals Networks (RPN) of Faster R-CNN\cite{ren2015faster}, which acts as extracting proposals and guiding the regression task of networks. Subsequently, the anchor mechanism within RPN is widely used in two-stage detectors\cite{jiang2017r2cnn,ma2018arbitrary,yang2019scrdet,cai2018cascade}. For one-stage detectors which detect objects\cite{redmon2016you,redmon2017yolo9000,liu2016ssd,lin2017focal} without RPN, YOLOv1\cite{redmon2016you} not use the anchor mechanism can't provide accuracy comparable to that of two-stage detectors. Afterwards, anchor methods are also extensively utilized in one-stage detectors\cite{redmon2017yolo9000,liu2016ssd,lin2017focal} to improve the performance of models. In the detection of oriented objects, most algorithms rely on anchor mechanism heavily. In general, these models output the oriented bounding boxes by rotating anchors to regress an additional angle information, and then obtain the final bounding boxes by the filtering of rotated NMS algorithm.  

The anchor mechanism promotes the development of object detection, but it is still not perfect and also has some problems like mentioned in \cite{law2018cornernet,zhou2019bottom,tian2019fcos,zhou2019objects}. Recently, the research of anchor-free has become a hot topic in the field of object detection . At present, the anchor-free detectors can be roughly divided into two categories. One is to locate objects through keypoints such as corner points in CornerNet\cite{law2018cornernet}, extreme points in ExtremeNet\cite{zhou2019bottom} and center points in CenterNet\cite{zhou2019objects}, the other is via the regression of a lot of points like FCOS\cite{tian2019fcos} to get the location of objects. For oriented objects, both of these two anchor-free categories have defects. In the inference stage, they all need to keep K of the highest scoring objects and may cause missed detection in the case of many targets in a single image like small cars in aerial image. $O^{2}$\textit{-DNet} is an one stage and anchor-free detector, which locates objects through a pair of median lines and their intersection point. In order to solve the top K problem, we combine the two categories of existing anchor-free alogrithm to design $O^{2}$\textit{-DNet}. The details of our model will be explained in the next section.

\section{O$^2$-DNet} \label{section:3}

\begin{figure*}[!t]
	\centering
	\includegraphics[width=12.5cm]{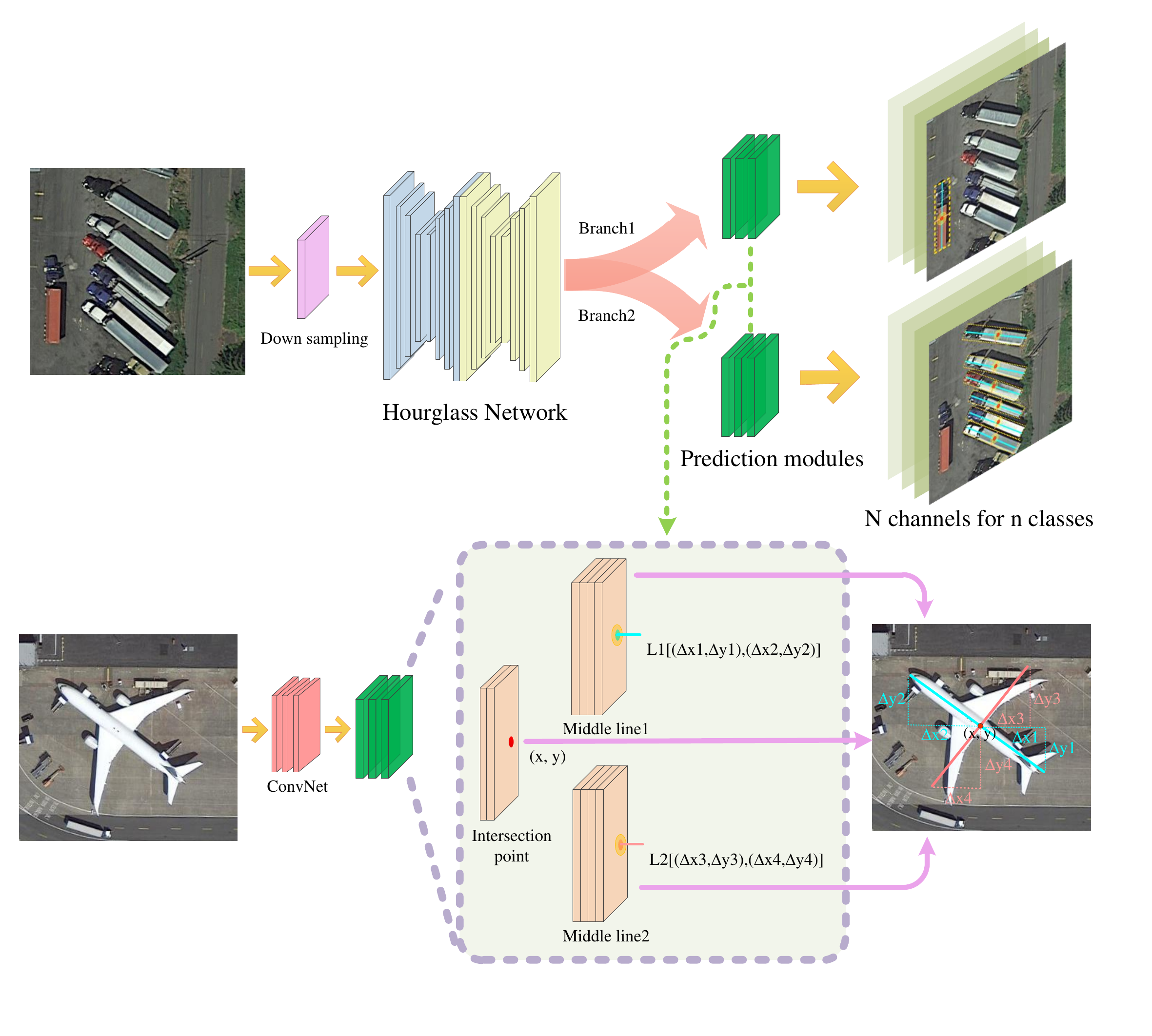}
	\caption{Architecture of $O^{2}$\textit{-DNet}. $O^{2}$\textit{-DNet} locates each object via a pair of median lines and their intersection point. Each middle line is represented by two corresponding endpoints. Two branches are added to detect horizontal and oriented objects respectively. $O^{2}$\textit{-DNet} outputs different classes of objects to different channels.} 
	\label{Figure 3}
\end{figure*}

\subsection{Overview}
Fig. \ref{Figure 3} illustrates the architecture of our method. $O^{2}$\textit{-DNet} locates per object by detecting a pair of  corresponding middle lines. We use 104-Hourgalss\cite{law2018cornernet} as the backbone of our model following the CornerNet \cite{law2018cornernet} for its excellent performance of extracting features. For an image of size $w \times h$, as input, our model outputs $2 \times (1 \times 2 \times C \times \frac{w}{d} \times \frac{h}{d})$ heatmaps to predict the intersection points of target middle lines, and $2 \times (2 \times 4 \times C \times \frac{w}{d} \times \frac{h}{d})$ regression maps to predict the corresponding two middle lines, where $C$ represents the number of classes in this image, the $d$ is output stride of down sampling module, and the first $2$ means two branches of $O^{2}$\textit{-DNet}. The design of two branches is to deal with the angle boundary problems of oriented objects via the independent prediction of objects with angle of 90 degree. For the prediction of middle lines, we obtain them by regressing their corresponding two endpoints. The form of regressing middle lines is inspired by CenterNet\cite{zhou2019objects}, it is the relative position of endpoint from intersection point. Moreover, we design special loss functions to control the relationship of endpoints to ensure the predicted median lines more accurate. In addition, in order to reduce the dependence of middle lines extraction on the precision of intersection point extraction, we propose the point drift region to make $O^{2}$\textit{-DNet} output high-quality oriented bounding boxes when the extraction of intersection points is not accurate enough.

\subsection{Hourglass Networks}\label{section:3.2}
Hourglass Networks\cite{newell2016stacked} was first proposed for human keypoints detection. In CornerNet\cite{law2018cornernet}, Law et al. modified hourglass network and introduced it into the field of object detection. We choose 104-Hourglass Networks modified in CornerNet\cite{law2018cornernet} as our backbone. For one image as input, HourglassNet regress $C$ channels of heatmaps with each pixel value $\check{y}\in(0, 1)$ which means the confidence of being judged as positive. Compared with CornerNet, $O^{2}$\textit{-DNet} defines each keypoint with value setting to 1, instead of the Gaussian Kernel. In the stage of inference, in order to avoid missed detection, instead of NMS and top K method to extract keypoints in heatmap used in \cite{law2018cornernet,zhou2019bottom,tian2019fcos,zhou2019objects}, we take a simple and rough way to extract keypoints, which is finding the connected domains in heatmap, and then define the center of each connected domain as the target keypoint. It is true that this method is not accurate enough, but it can achieve satisfactory results in the experiments through the collocation with intersection drift region proposed in our model.

\begin{figure*}[!t]
	\centering
	\includegraphics[width=12.5cm]{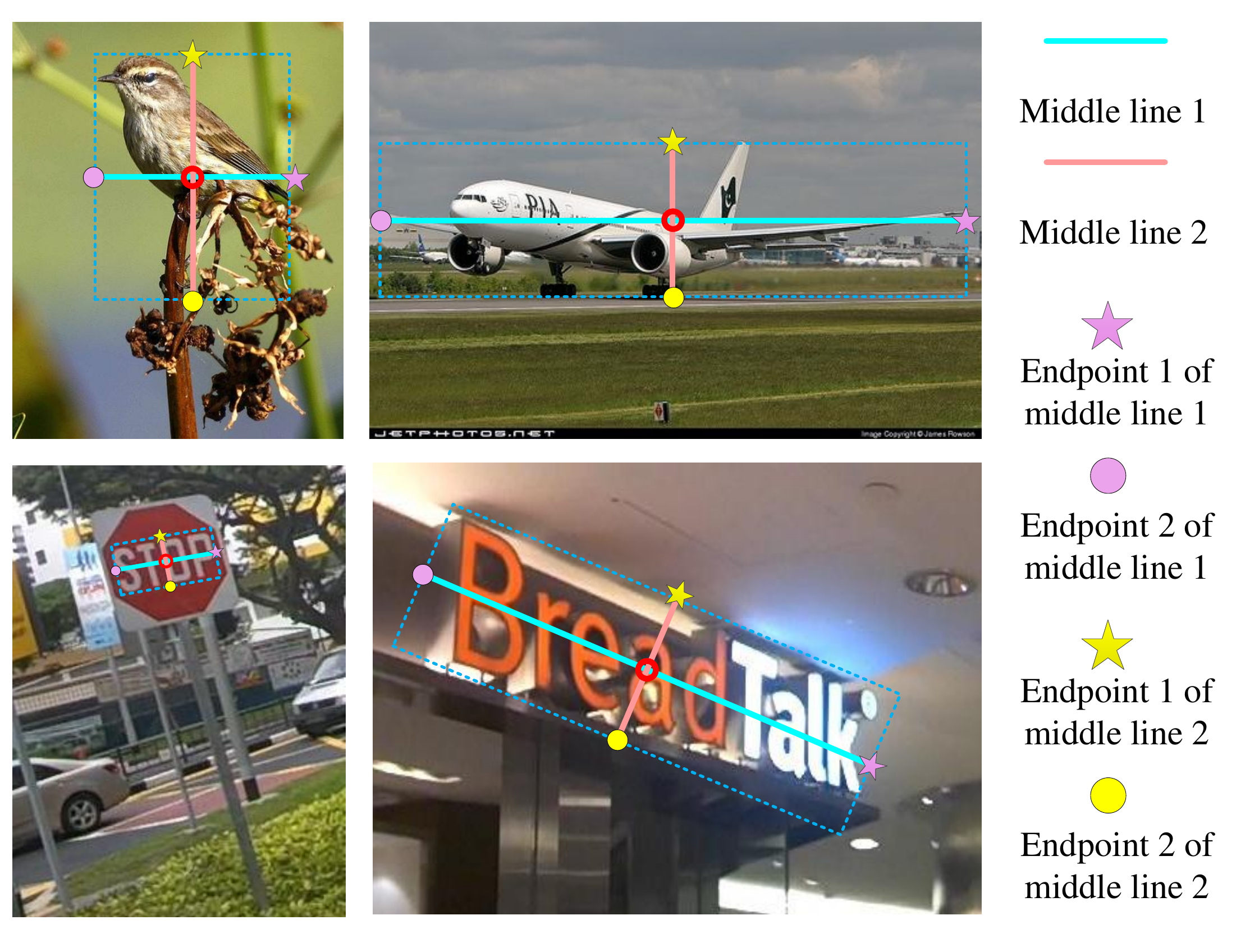}
	\caption{The definition of two middle lines and the order of two endpoints of per middle line. The first line of figures represent objects with angles of 90 degree, and the second line of figures represent oriented objects with another angles.} 
	\label{Figure 4}
\end{figure*}

\subsection{Middle Lines and Their Intersection Point} \label{section:3.3}

As shown in Fig. \ref{Figure 4}, let $L_1[(x1,y1),(x2,y2)]$ denote middle line 1 of the object, $(x1,y1)$ and $(x2,y2)$ are the endpoint 1 and 2 of $L_1$ respectively. Similarly, $L_2[(x3,y3),(x4,y4)]$ is defined as the middle line 2. For the first branch of $O^{2}$\textit{-DNet}, we define the horizontal median line as $L_1$, and the vertical one as $L_2$. We regard the right endpoint and the top one as the endpoint 1 of $L_1$ and $L_2$ respectively. For the second branch, the longer one of the two median lines is defined as $L_1$, the other is $L_2$. Endpoints 1 are also defined as the right one and the top one in $L_1$ and $L_2$ in this branch. The intersection point $(x,y)$ of two middle lines can be obtained through simple operators $(\frac{x1+x2+x3+x4}{4},\frac{y1+y2+y3+y4}{4})$ in both two branches.

\subsubsection{Intersection Point}

We follow the modified focal loss\cite{lin2017focal} in CornerNet\cite{law2018cornernet} to predict the heatmap of intersection point of target middle lines. Because the ground truth of our model with value setting to 1 instead of the Gaussian Kernel like mentioned in {\color{red}\textit{Section \ref{section:3.2}}}, the loss in $O^{2}$\textit{-DNet} is a little different compared with CornerNet in form. We name the loss of intersection point in our model as $\mathcal{L}_{ip}$:

\begin{eqnarray}
\mathcal{L}_{ip} = -\frac{1}{N}\sum_{xy}
\begin{cases}
(1-\acute{{Y_{xy}}})^\alpha\log \acute{{Y_{xy}}},  & \text{\ if $Y_{xy}$ = 1} \\
\acute{{Y_{xy}}}^\alpha\log(1-\acute{{Y_{xy}}}), & \text{\ if $Y_{xy}$ = 0}
\end{cases}
\vspace{2 ex}
\end{eqnarray}
where $\alpha$ is a hyper-parameter and the value fixed to 2 in our experiment. $\acute{Y_{xy}}$ represents the pixel value at the coordinate $(x,y)$ in heatmap and $Y_{xy}$ corresponds to the ground truth. $N$ is the number of objects.

\begin{figure*}[!t]
	\centering
	\includegraphics[width=12.5cm]{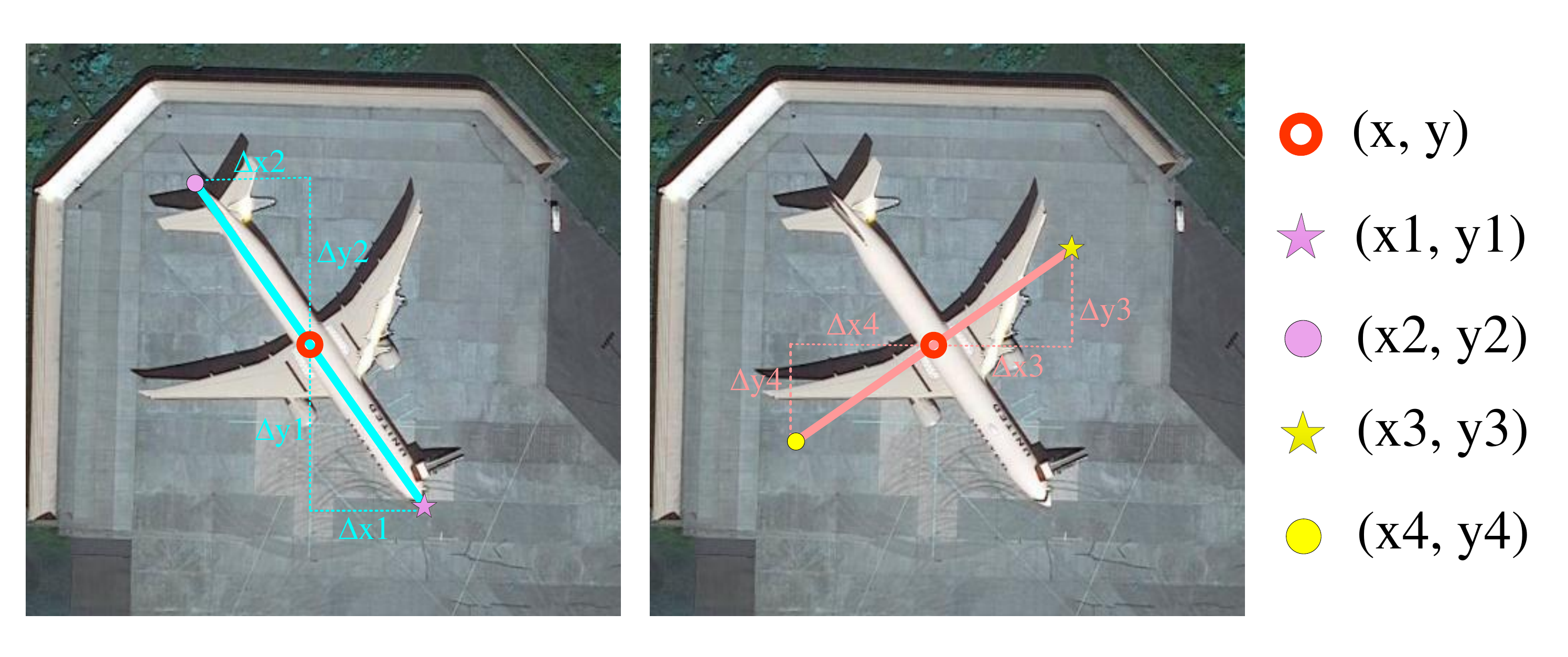}
	\caption{The method of regressing middle lines in our model. The relative distance between endpoint and intersection point is represented as $\Delta$ xi, $\Delta$ yi, where i is 1,2,3,4 meaning the four corresponding endpoints of middle lines.} 
	\label{Figure 5}
\end{figure*}

\subsubsection{Middle Lines} \label{section:3.3.2}

The method of regressing middle lines is to regress the relative distance between each endpoint of per middle line with the intersection point. As shown in Fig. \ref{Figure 5}, for middle line $1$, we need to regress $4 \times H \times W$ maps, and the values of these 4 maps in the position of intersection point are $\Delta x1, \Delta y1, \Delta x2$ and $\Delta y2$ respectively. The form of regression of middle line $2$ is the same as middle line $1$. The loss to regress each middle line is as follows:

\begin{eqnarray}
\mathcal{L}_{1} = \frac{1}{N}\sum_{ep=1}^2\sum_{\hat{xy}}[SmoothL1(\Delta x^*_{\hat{xy}}, \Delta x_{\hat{xy}}) + SmoothL1(\Delta y^*_{\hat{xy}}, \Delta y_{\hat{xy}})]_{ep}
\end{eqnarray}
where $N$ is the number of objects. $ep$ denotes the endpoint of the corresponding middle line. $\hat{xy}$ means the coordinate of the regression map. $\Delta x^*$ and $\Delta y^*$ represent the ground truth. 

The way of regressing each endpoint of per middle line independently may result in two endpoints and the intersection point being not collinear. In order to address this problem, we introduce a loss function as follows:

\begin{eqnarray}
\mathcal{L}_{2} = \frac{1}{N}\sum_{l=1}^2\sum_{\hat{xy}}[SmoothL1(\Delta x^{ep1}_{\hat{xy}} \times \Delta y^{ep2}_{\hat{xy}}, \Delta x^{ep2}_{\hat{xy}} \times \Delta y^{ep1}_{\hat{xy}})]_{l}
\end{eqnarray}
where $l$ means two middle lines of per object. $ep1$ and $ep2$ denote the endpoint $1$ and $2$ of each middle line.

There are two middle lines of one object, and they are vertical in space generally. In order to control the relationship of two target middle lines, we design $\mathcal{L}_{3}$ as follows:

\begin{eqnarray}
\mathcal{L}_{3} = \frac{1}{N}\sum_{\hat{xy}}[SmoothL1(\Delta x^{ep1_{l1}}_{\hat{xy}} \times \Delta x^{ep1_{l2}}_{\hat{xy}} + \Delta y^{ep1_{l1}}_{\hat{xy}} \times \Delta y^{ep1_{l2}}_{\hat{xy}}, 0)]_{l}
\end{eqnarray}
where $ep1_{l1}$ means endpoint $1$ of middle line $1$, $ep1_{l2}$ means endpoint $1$ of middle line $2$.

The $\mathcal{L}_{1}$, $\mathcal{L}_{2}$ and $\mathcal{L}_{3}$ make up the Line Loss of our model:
\begin{eqnarray}
\mathcal{L}_{l} = \mathcal{L}_{1} + \alpha\mathcal{L}_{2} + \beta\mathcal{L}_{3}
\end{eqnarray}

And the total loss of our model can be expressed as:
\begin{eqnarray}
\mathcal{L}_{oss} = \mathcal{L}_{ip} + \gamma\mathcal{L}_{l}
\end{eqnarray}
where  $\alpha$, $\beta$ and $\gamma$ are weights of losses.

\begin{figure*}[!t]
	\centering
	\includegraphics[width=12.5cm]{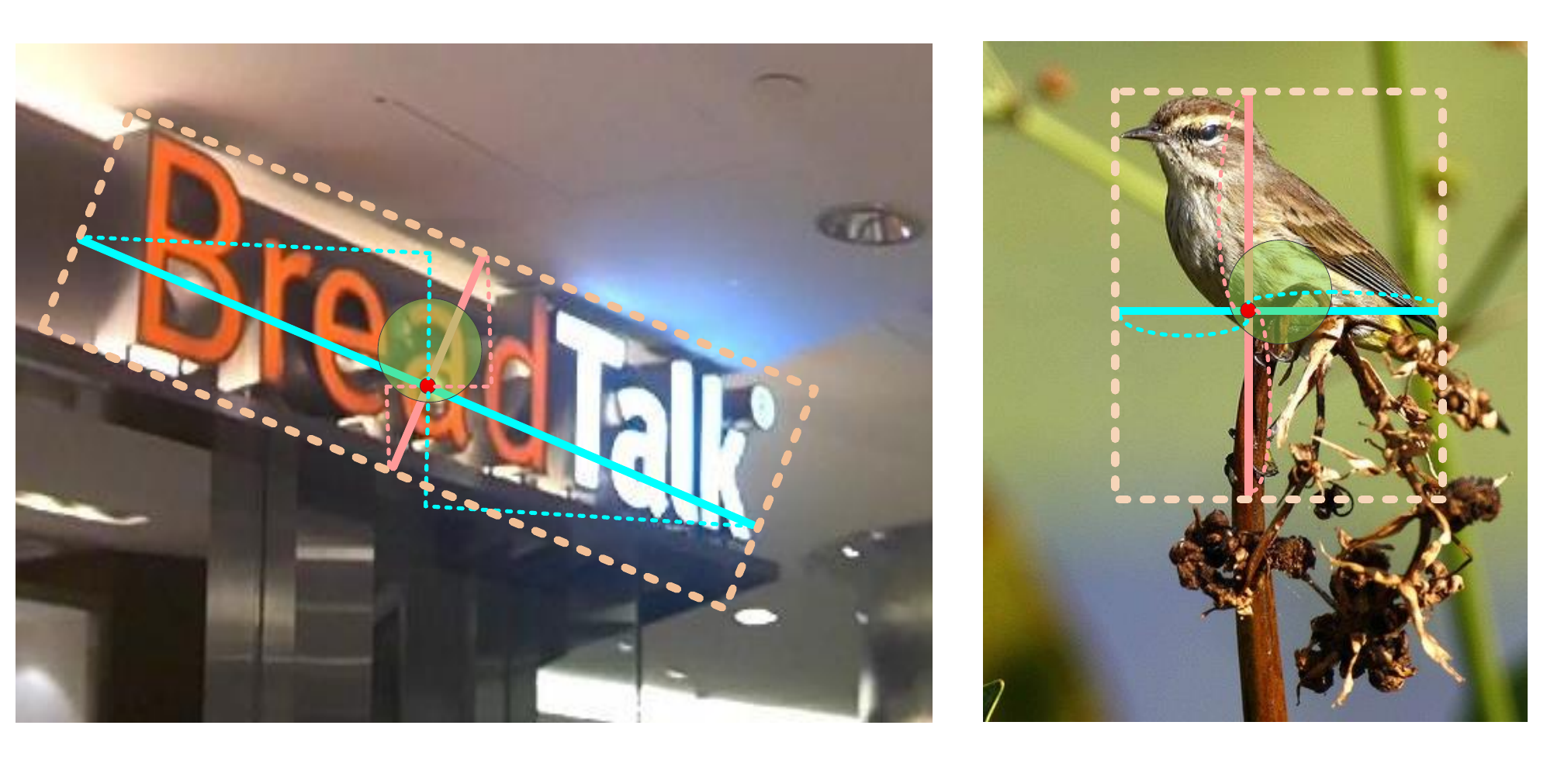}
	\caption{Illustration of drift region. Per pixel in drift region will regress the relative distances from 4 endpoints of corresponding two middle lines.} 
	\label{Figure 5}
\end{figure*}

\subsubsection{Drift Region}

The extraction of intersection point in heatmap will affect the accuracy of middle lines extraction. Inspired by FCOS\cite{tian2019fcos}, we set up circular drift regions in the center of objects according to the size of them. All pixel points in drift region will regress the different values according to their relative distances from endpoints of middle lines. The drift region guarantees that the extraction of intersection point from heatmap will not influence the position of final oriented bounding box. The radius of the drift region is set as follows:

\begin{eqnarray}
\mathcal{R} = \min[r / stride, \min(L_{1}, L_{2}) / (2 \times stride)]
\end{eqnarray}
where stride is output stride of our model, $L_{1}$ and $L_{2}$ are the middle line $1$ and $2$ respectively. Where $r$ is $16$ in our model. 

Unlike remote sensing images, objects in natural scenes sometimes have overlaps to form fuzzy samples, we also follow FCOS to address this problem which is when a pixel belongs to two targets at the same time, we regress the small one.

\section{Experiments} \label{section:4}

\subsection{Datasets}
In the stage of experiments, we select three datasets to verify the performance of our model. These datasets involve in different research fields: oriented objects detection of aerial images, text detection in natural scene, the detection of objects in nature images. Their detailed introductions are as follows: 

\subsubsection{DOTA}

DOTA\cite{Xia_2018_CVPR} is a common benchmark for the detection of objects in aerial images. It includes two detection tasks: horizontal bounding boxes and oriented bounding boxes, and we only use the oriented one in our experiments. There are $2806$ aerial images with size ranges from $800 \times 800$ to $4000 \times 4000$ pixels total in DOTA. These images are annotated using $15$ categories (e.g., aircraft, small car, ship). In practice, we divide each large image to crop images in $800 \times 800$ with overlap of $0.25$. 

\subsubsection{ICDAR 2015}
ICDAR 2015\cite{karatzas2015icdar} is a dataset used for the detection of text in natural scene. The training set and test set include $1000$ and $500$ images with the size of $720 \times 1280$, respectively.

\subsubsection{COCO}
The challenging MS COCO\cite{lin2014microsoft} dataset contains 80k images for training, 40k for validation and 20k for testing and includes 80 categories. The annotations in COCO are horizontal bounding boxes which we used to test the generality of our model.

\subsection{Training and Testing Details}

All our experiments are implemented on PyTorch 1.0\cite{paszke2017automatic} by two NVIDIA Tesla V100 GPUs with $2 \times 32$ GB memories. For DOTA, we set the input resolution $800 \times 800$ to $511 \times 511$ and the output stride to $4$ following settings in \textit{CornerNet}\cite{law2018cornernet} during the training stage. Adam\cite{kingma2014adam} is selected as the optimizer for \textit{$O^{2}$\textit{-DNet}}. We train our model from scratch to $300k$ iterations with the batch size setting to $32$. The learning rate starts from 0.001 and 10 times lower for every third iterations. Simple random horizontal and vertical flipping as well as color dithering are used to enhance the data in our experiments. The weights of loss $\alpha$, $\beta$ and $\gamma$ ({\color{red}\textit{Section \ref{section:3.3.2}}}) are setting to $1,1$ and $0.5$ respectively during training.  For ICDAR 2015 and COCO, \textit{$O^{2}$\textit{-DNet}} is finetuned on two v100 GPUs for $200k$ iterations with a batch size of $32$ from a pre-trained CornerNet model which trained on 10 GPUs for 500k iterations. Other settings are the same as DOTA. It is worth noting that for the two branches of our model, we do not strictly divide them by $90$ degrees, but by an angle range of $(88, 92)$ degrees.

During test stage, as mentioned in {\color{red}\textit{Section \ref{section:3.2}}}, we need to transform the heatmap into a binary image to extract the intersection point of two target median lines, where the threshold is set to $0.3$. When the angle of an object is critical in two branches, it may have output in both two. We take the one with the highest intersection point score as the final output of \textit{$O^{2}$\textit{-DNet}}.

\subsection{Comparisons with State-of-the-art Frameworks}

In this part, we first prove the advancement of \textit{$O^{2}$\textit{-DNet}} on the oriented objects datasets (DOTA, ICDAR 2015). Then we test the strong generality of our model on the  dataset of natural objects (COCO).

\subsubsection{DOTA}

As shown in Table \ref{table:1}, our O$^2$DNet achieve $71.04\%$ mAP on DOTA dataset, better than most two-stage and one-stage models used in the detection of aerial objects at present. For bridges with large aspect ratio and dense parked small vehicles, our anchor-free model achieves the most advanced accuracy on AP due to the better adaptive feature extraction ability than anchor-based models.

\begin{table*}[!t]
	\centering
	\resizebox{1.0\textwidth}{!}{
		\begin{tabular}{l|c|c|c|c|c|c|c|c|c|c|c|c|c|c|c|c}
			\hline
			\hline
			Method &  Pl &  Bd &  Br &  Gft &  Sv &  Lv &  Sh &  Tc &  Bc &  St &  Sbf &  Ra &  Ha &  Sp &  He &  mAP\\
			\hline
			\hline
			
			\textbf{Two-stage models} & \multicolumn{16}{c}{} \\
			\hline
		
			R$^2$CNN \cite{jiang2017r2cnn} & 80.94 & 65.67 & 35.34 & 67.44 & 59.92 & 50.91 & 55.81 & 90.67 & 66.92 & 72.39 & 55.06 & 52.23 & 55.14 & 53.35 & 48.22 & 60.67 \\
			RRPN \cite{ma2018arbitrary} & 88.52 & 71.20 & 31.66 & 59.30 & 51.85 & 56.19 & 57.25 & 90.81 & 72.84 & 67.38 & 56.69 & 52.84 & 53.08 & 51.94 & 53.58 & 61.01 \\
			
			R-DFPN \cite{yang2018automatic} & 80.92 & 65.82 & 33.77 & 58.94 & 55.77 & 50.94 & 54.78 & 90.33 & 66.34 & 68.66 & 48.73 & 51.76 & 55.10 & 51.32 & 35.88 & 57.94 \\

			ICN \cite{azimi2018towards} & 81.40 & 74.30 & 47.70 & 70.30 & 64.90 & 67.80 & 70.00 & 90.80 & 79.10 & 78.20 & 53.60 & 62.90 & 67.00 & 64.20 & 50.20 & 68.20 \\
			
			RoI-Transformer \cite{Ding_2019_CVPR} & 88.64 & 78.52 & 43.44 & {\bf 75.92} & 68.81 & 73.68 & {\bf 83.59} & 90.74 & 77.27 & 81.46 & 58.39 & 53.54 & 62.83 & 58.93 & 47.67 & 69.56 \\
			
			SCRDet \cite{yang2019scrdet} & {\bf89.98} & 80.65 & {\bf52.09} & 68.36 & 68.36 & 60.32 & 72.41 & {\bf90.85} & {\bf87.94} & {\bf86.86} & 65.02 & {\bf66.68} & 66.25 & 68.24 & {\bf65.21} & {\bf72.61}\\
			\hline
			\textbf{One-stage models} & \multicolumn{16}{|c}{} \\
			\hline
			SSD \cite{liu2016ssd} & 39.83 & 9.09 & 0.64 & 13.18 & 0.26 & 0.39 & 1.11 & 16.24 & 27.57 & 9.23 & 27.16 & 9.09 & 3.03 & 1.05 & 1.01 & 10.59 \\
			YOLOv2 \cite{redmon2016you} & 39.57 & 20.29 & 36.58 & 23.42 & 8.85 & 2.09 & 4.82 & 44.34 & 38.35 & 34.65 & 16.02 & 37.62 & 47.23 & 25.5 & 7.45 & 21.39 \\
			RetinaNet \cite{lin2017focal} & 88.92 & 67.67 & 33.55 & 56.83 & 66.11 & 73.28 & 75.24 & 90.87 & 73.95 & 75.07 & 43.77 & 56.72 & 51.05 & 55.86 & 21.46 & 62.02 \\
			R$^3$Det \cite{yang2019r3det} & 89.54 & 81.99 & 48.46 & 62.52 & 70.48 & 74.29 & 77.54 & 90.80 & 81.39 & 83.54 & 61.97 & 59.82 & 65.44 & 67.46 & 60.05 & 71.69\\
			O$^2$-DNet  & 89.31 & {\bf82.14} & 47.33 & 61.21 & {\bf71.32} & 74.03 & 78.62 & 90.76 & 82.23 & 81.36 & 60.93 & 60.17 & 58.21 & 66.98 & 61.03 & 71.04\\
			\hline
	\end{tabular}}
	\vspace{10pt}
	
	\caption{State-of-the-art comparisons on DOTA.  The abbreviations of the names are defined as: Pl: Plane, Bd: Baseball diamond, Br: Bridge, Gft: Ground field track, Sv: Small vehicle, Lv: Large vehicle, Sh:Ship, Tc: Tennis court, Bc: Basketball court, St: Storage tank, Sbf: Soccer-ball field, Ra: Roundabout, Ha: Harbor, Sp: Swimming pool, and He: Helicopter. The SSD, YOLOv2 and RetinaNet are modified by us to output oriented bounding boxes.}
	\label{table:1}
\end{table*}

\subsubsection{ICDAR 2015}

For ICDAR 2015 dataset, most of the annotation of objects is not in the form of rectangle, but in the form of irregular quadrilateral which is close to parallelogram. We shield the $\mathcal{L}_{3}$ of the Line Loss, which is used to control the two target middle line to remain vertical. As shown in Table \ref{table:2}, our O$^2$-DNet achieve $82.97\%$ F1, better than other models we choose for comparison. The experimental results show that our model can be used not only in the detection of aerial images but also in natural scene text detection.

\begin{table}[!h]
\centering
\resizebox{1\textwidth}{!}{
	\begin{tabular}{l|l|l|c| |l|l|l|c}
		\hline
		\hline
		Method & Recall & Precision & F1 & Method & Recall & Precision & F1\\
		\hline
		\hline
		CTPN \cite{tian2016detecting} & 51.56 & 74.22 & 60.85 & SegLink \cite{shi2017detecting} & 76.80 & 73.10 & 75.00   \\
		R$^2$CNN \cite{jiang2017r2cnn}  & 79.68 & \bf{85.62} & 82.54  & EAST \cite{Zhou2017EAST} & 78.33 & 83.27 & 80.72\\
		FOTS RT \cite{liu2018fots} & \bf{85.95} & 79.83 & 82.78& RRPN \cite{ma2018arbitrary} & 82.17 & 73.23 & 77.44 \\
		\hline
		\hline
		O$^2$-DNet  & 80.52 & 85.58 & \bf{82.97} \\
		\hline		
	
\end{tabular}}
\vspace{10pt}
\caption{Comparison with different methods on the ICDAR2015 dataset. Our model achieve $82.97\%$ F1, better than other models in this table.}
\label{table:2}
\end{table}

\subsubsection{COCO}

In order to verify the general performance of our model, we also make experiments on the COCO dataset of natural scene object detection. For COCO with objects labeled in horizontal bounding boxes, our model will only have output in the first branch. As shown in Table \ref{table:3}, our O$^2$-DNet achieve $41.3\%$ AP on COCO dataset, leading most one-stage detectors.

\begin{table*}[!h]
	\begin{center}
		\resizebox{1\textwidth}{!}{
			\begin{tabular}{l|c|c|c|c|c|c|c}
				\toprule
				\toprule
				Method& Backbone  & $AP$ & $AP_{50}$ & $AP_{75}$ & $AP_{S}$ & $AP_{M}$ & $AP_{L}$ \\
				\midrule
				\midrule
				\multicolumn{8}{l}{\textbf{Two-stage detectors}} \\
				\midrule
				Faster R-CNN\cite{lin2017feature} & ResNet-101-FPN  & 36.2 &
				59.1 & 39.0 & 18.2 & 39.0 & 48.2\\ 
				Deformable-CNN\cite{dai2017deformable} & Inception-ResNet& 37.5 &
				58.0 & - & 19.4 & 40.1 & 52.5 \\
				Mask R-CNN\cite{he2017mask} & ResNeXt-101  & 39.8 & 62.3 & 43.4 &
				22.1 & 43.2 & 51.2 \\
				Cascade R-CNN\cite{cai2018cascade} & ResNet-101  & 42.8 & 62.1 &
				46.3 & 23.7 & 45.5 & 55.2\\	
				D-RFCN + SNIP\cite{singh2018analysis} & DPN-98
				& 45.7 & 67.3 & 51.1 & 29.3 & 48.8 & 57.1 \\
				PANet\cite{liu2018path} & ResNeXt-101 
				& \bf{47.4} & \bf{67.2} & \bf{51.8} & \bf{30.1} & \bf{51.7} & \bf{60.0} \\ 
				\midrule
				\multicolumn{8}{l}{\textbf{One-stage detectors}} \\
				\midrule
				YOLOv2\cite{redmon2017yolo9000} & DarkNet-19 & 21.6 & 44.0 & 19.2
				& 5.0 & 22.4 & 35.5 \\	
				SSD\cite{liu2016ssd} & ResNet-101 & 31.2 & 50.4 & 33.3 & 10.2 &
				34.5 & 49.8 \\
				RetinaNet\cite{lin2017focal} & ResNet-101 & 39.1 & 59.1 & 42.3
				& 21.8 & 42.7 & 50.2 \\
				RefineDet\cite{zhang2018single} & ResNet-101 &
				36.4 & 57.5 & 39.5 & 16.6 & 39.9 & 51.4 \\
				CornerNet\cite{law2018cornernet} & 104-Hourglass & 40.5 & 56.5 & 43.1 &
				19.4 & 42.7 & 53.9 \\
				ExtremeNet\cite{zhou2019bottom}& 104-Hourglass & 40.2 & 55.5 & 43.2 & 20.4 & 43.2 & 53.1 \\
				CenterNet\cite{zhou2019objects}& 104-Hourglass & \bf{42.1} & \bf{61.1} & \bf{45.9} & 24.1 & \bf{45.5} & \bf{52.8} \\
				FCOS\cite{tian2019fcos}& ResNet-101-FPN & 41.0 & 60.7 & 44.1 & 24.0 & 44.1 & 51.0\\ 
				O$^2$-DNet& 104-Hourglass & 41.3 & 60.9 & 45.2 & \bf{24.2} & 44.5 & 52.3
				\\
				\bottomrule
			\end{tabular}}
		\vspace{10pt}
		\caption{State-of-the-art comparison on COCO test-dev. It shows that our O$^2$-DNet has a strong competitiveness in natural scene object detection. In this table, all data are from single scale detection results.}
		\label{table:3}
	\end{center}
\end{table*}

\begin{table*}[!h]
	\centering
	\resizebox{1.0\textwidth}{!}{
		\begin{tabular}{l|c|c|c|c|c|c|c|c|c|c|c|c|c|c|c|c}
			\hline
			\hline
			Method &  Pl &  Bd &  Br &  Gft &  Sv &  Lv &  Sh &  Tc &  Bc &  St &  Sbf &  Ra &  Ha &  Sp &  He &  mAP\\
			\hline
			\hline
			
			ResNet101-FPN\cite{lin2017feature}  & 88.07 & 81.43 & 47.03 & 60.51 & 68.42 & 70.96 & 73.72 & 90.01 & 81.15 & 78.24 & 59.83 & 57.52 & 56.01 & 62.63 & 58.37 & 68.93\\
			Without Line Loss  & 87.25 & 82.12 & 47.04 & 60.14 & 67.21 & 70.01 & 71.38 & 90.26 & 79.89 & 81.09 & 58.43 & 59.12 & 56.05 & 66.82 & 60.06 & 69.12\\
			Single branch  & 86.17 & 81.07 & 47.06 & 60.28 & 66.28 & 72.44 & 69.09 & 88.90 & 81.29& 78.77 & 59.12 & 57.97 & 58.14 & 64.69 & 59.98 & 68.81\\
			O$^2$-DNet  & 89.31 & 82.14 & 47.33 & 61.21 & 71.32 & 74.03 & 78.62 & 90.76 & 82.23 & 81.36 & 60.93 & 60.17 & 58.21 & 66.98 & 61.03 & 71.04\\
			\hline
	\end{tabular}}
	\vspace{10pt}	
	\caption{Ablation Studies on DOTA. \textit{ResNet101-FPN} means that the backbone of our model is replaced by ResNet101-FPN. \textit{Without Line Loss} denotes that we only keep the $\mathcal{L}_{1}$ part of the total Line Loss. \textit{Single branch} means that we only keep the second branch of O$^2$-DNet.}
	\label{table:4}
\end{table*}

\subsection{Ablation Studies}

In this part, we conduct three ablation experiments on the DOTA dataset, which are the influence of different backbones on the performance of our model, the influence of Line Loss on the performance of our model, and the influence of single branch on the performance of our model. Table \ref{table:4} shows all experimental data. The following is the specific analysis:

\subsubsection{ResNet101-FPN} 

We replace the backbone 104-Hourglass with ResNet101-FPN\cite{lin2017feature} to verify the effect of our model with different backbones. As shown in Table \ref{table:4}, our model  also achieve satisfactory $68.93\%$ mAP in matching with ResNet101-FPN, which means that the performance of O$^2$-DNet does not depend entirely on 104-Hourglass.

\subsubsection{Without Line Loss}

\begin{figure*}[!h]
	\centering
	\includegraphics[width=12.2cm]{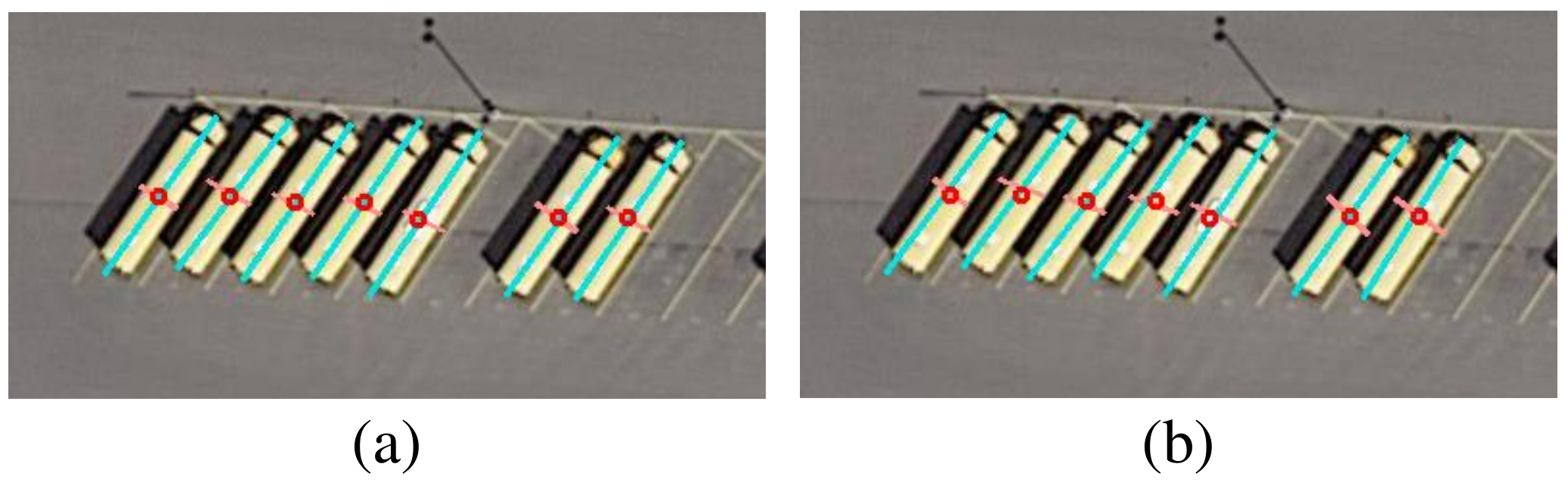}
	\caption{Effects of Line Loss. Figure (a) shows the output of our model with Line Loss and Figure (b) shows the result of our model without Line Loss.} 
	\label{Figure 6}
\end{figure*}

In order to prove the validity of Line Loss, we shield the $\mathcal{L}_{2}$, $\mathcal{L}_{3}$ part of Line Loss, and keep the other settings of O$^2$-DNet. Table \ref{table:4} shows that our model with Line Loss improves $1.92\%$ mAP compared with the model without Line Loss. The Line Loss effectively controls the line segment property of the regression target median lines in our model. The effect of Line Loss is shown in Fig. \ref{Figure 6}.

\subsubsection{Single branch}

In order to verify that the two branches of O$^2$-DNet can solve the boundary problem better, we cut off the first branch and input the 90 degree object into the second branch in the form of the original ground truth defined in {\color{red}\textit{Section \ref{section:3.3}}}. The experimental results show that the mAP of two branches is $2.23\%$ higher than that of single branch. The design of two branches is significant for our model.

\begin{figure*}[!h]
	\centering
	\includegraphics[width=12.5cm]{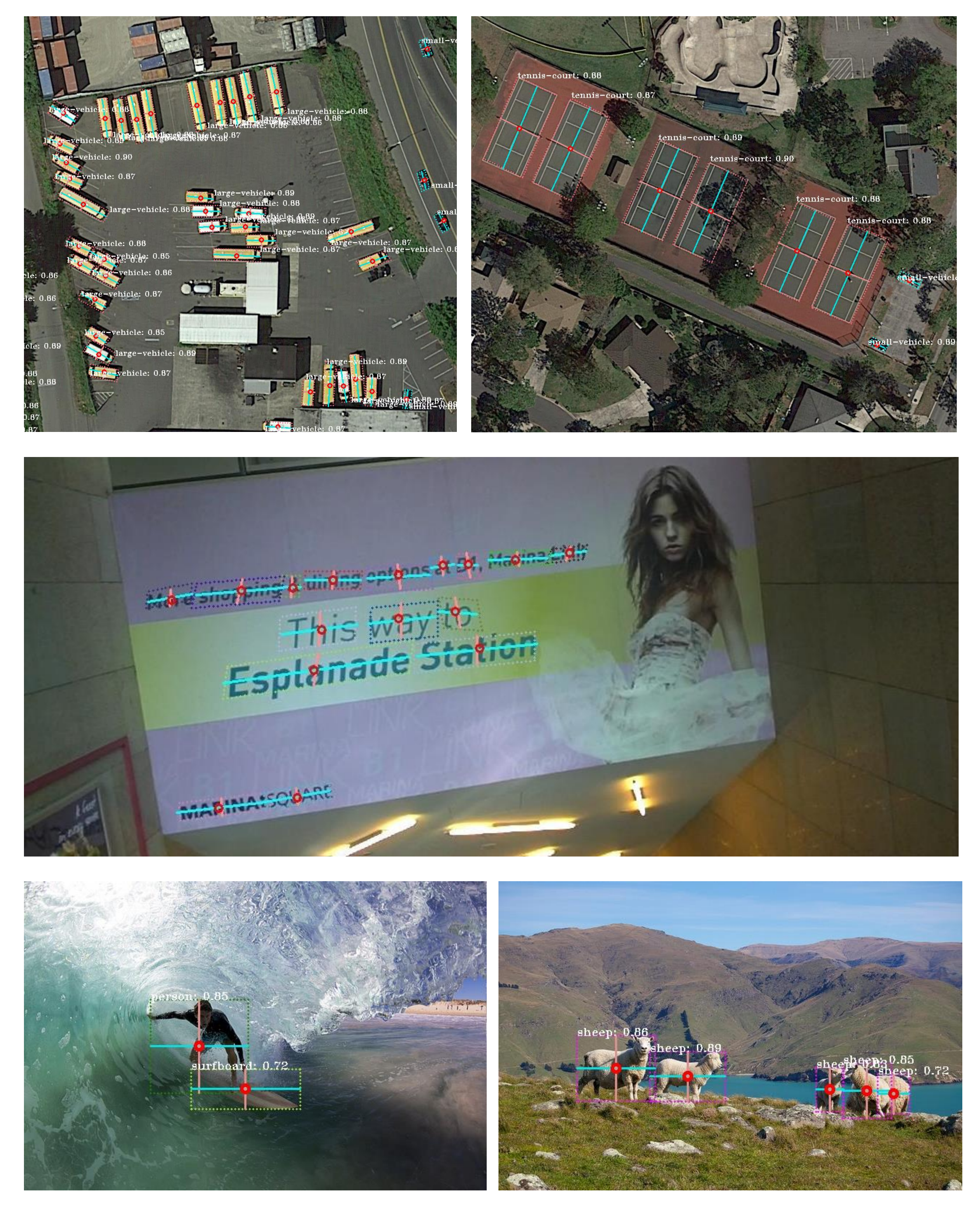}
	\caption{Qualitative results output by O$^2$-DNet.} 
	\label{Figure 7}
\end{figure*}

\section{Conclusion} \label{section:5}

We propose a novel one-stage and anchor-free model named $O^{2}$\textit{-DNet} to detect oriented objects. $O^{2}$\textit{-DNet} locates each object by predicting a pair of middle lines inside them. As a result, our model is competitive compared with state-of-the-art detectors in several fields: oriented objects detection of aerial images, text detection in natural scene, the detection of objects in nature images.

\bibliographystyle{splncs}
\bibliography{eccv}

\end{document}